\documentclass[a4paper]{article}
\usepackage{graphicx}
\usepackage{amssymb} 
\usepackage{amsmath}
\usepackage{subcaption}
\usepackage{float}

\begin{document}

	\title{\Large Semi--supervised Graph Embedding Approach to Dynamic Link Prediction}
	\author{Ryohei Hisano\thanks{Social ICT Research Center, Graduate School of Information Science and Technology, The University of Tokyo, email: em072010@yahoo.co.jp}}
	\date{}
	
	\maketitle

	
	\begin{abstract} \small\baselineskip=9pt We propose a simple discrete time semi--supervised graph embedding approach to link prediction in dynamic networks.  The learned embedding reflects information from both the temporal and cross--sectional network structures, which is performed by defining the loss function as a weighted sum of the supervised loss from past dynamics and the unsupervised loss of predicting the neighborhood context in the current network.  Our model is also capable of learning different embeddings for both formation and dissolution dynamics.  These key aspects contributes to the predictive performance of our model and we provide experiments with three real--world dynamic networks showing that our method is comparable to state of the art methods in link formation prediction and outperforms state of the art baseline methods in link dissolution prediction.\end{abstract}

	\section{Introduction}
	
	One of the central tasks concerning network data is the problem of link prediction.  Link prediction can be roughly divided into two types: static link prediction and temporal link prediction.  Static link prediction is concerned with the problem of predicting the overall structure of a network.  The goal is to predict missing links in partially observed network data that are absent from the dataset but that should in fact exist.  Example applications include knowledge graph completion, predicting relationships among participants in social networking services and protein-protein interactions.  We refer to \cite{Liben-Nowell2003,Getoor2005,Clauset2008} for excellent reviews of the field.  In a temporal link prediction problem, the goal is to predict the future network state given previous linkage patterns \cite{Sarkar2007,Hasan2006,Dunlavy2011}.  Example applications include recommender systems where users and products are modeled as a bipartite graph and user purchases are modeled as linkages over time.  The goal here is to predict future purchase patterns of users from past purchase patterns.

	In this paper, we focus on a slight variation of the temporal link prediction problem.  Given a sequence of network snapshots from time $1$ to time $t$, our problem is to predict the \textit{transition} of a network from time $t$ to time $t+1$.  A \textit{transition} of a network can be summarized using two networks, a link formation network and a link dissolution network.  We choose to predict the \textit{transition} of a network instead of a network at the next time step for three main reasons.  Firstly, by predicting a network only at the next time step, one cannot distinguish whether the prediction of link formation is successful, whether the prediction of link dissolution is successful or whether the network itself did not change much between different time steps, and whether simply using the network information from the last time step might suffice for prediction.  We want to avoid this redundancy by focusing on predicting the \textit{transition}.  Secondly, different forces might govern link formation and link dissolution.  Our hope is that by separately modeling these forces we might obtain better predictive accuracy.  Thirdly, predicting link dissolution is important in its own right.  For instance, in the financial crisis of 2008, many banks were reported to dissolve their relationships with poorly performing firms while forming new links with better performing firms.  Being able to predict the formation and dissolution dynamics of a network separately in this setting is an important issue in risk management.  This is true even in social networks, where important dissolutions in links might prevent the spread of good or bad influences in a community \cite{Christakis2007}.

	Our modelling approach is a variant of semi--supervised graph embedding \cite{Yang2016}.  The supervised part consists of a complex--valued latent feature bilinear model \cite{Trouillon2016} where past link formation and link dissolution information plays the role of target values in the training data.  The unsupervised part consists of a graph embedding predicting the neighborhood context in the current network \cite{Perozzi2014}.  The same complex--valued vectors are used in both tasks, and the weighted sum of these two losses is the total loss in our model.  Semi--supervised graph embedding \cite{Yang2016} was originally intended for use in node classification, but we extend the idea to learning complex--valued vectors capable of predicting the \textit{transition} of a network.

\begin{figure*}[!h]
		\begin{center} 
			\includegraphics*[width=.7\textwidth]{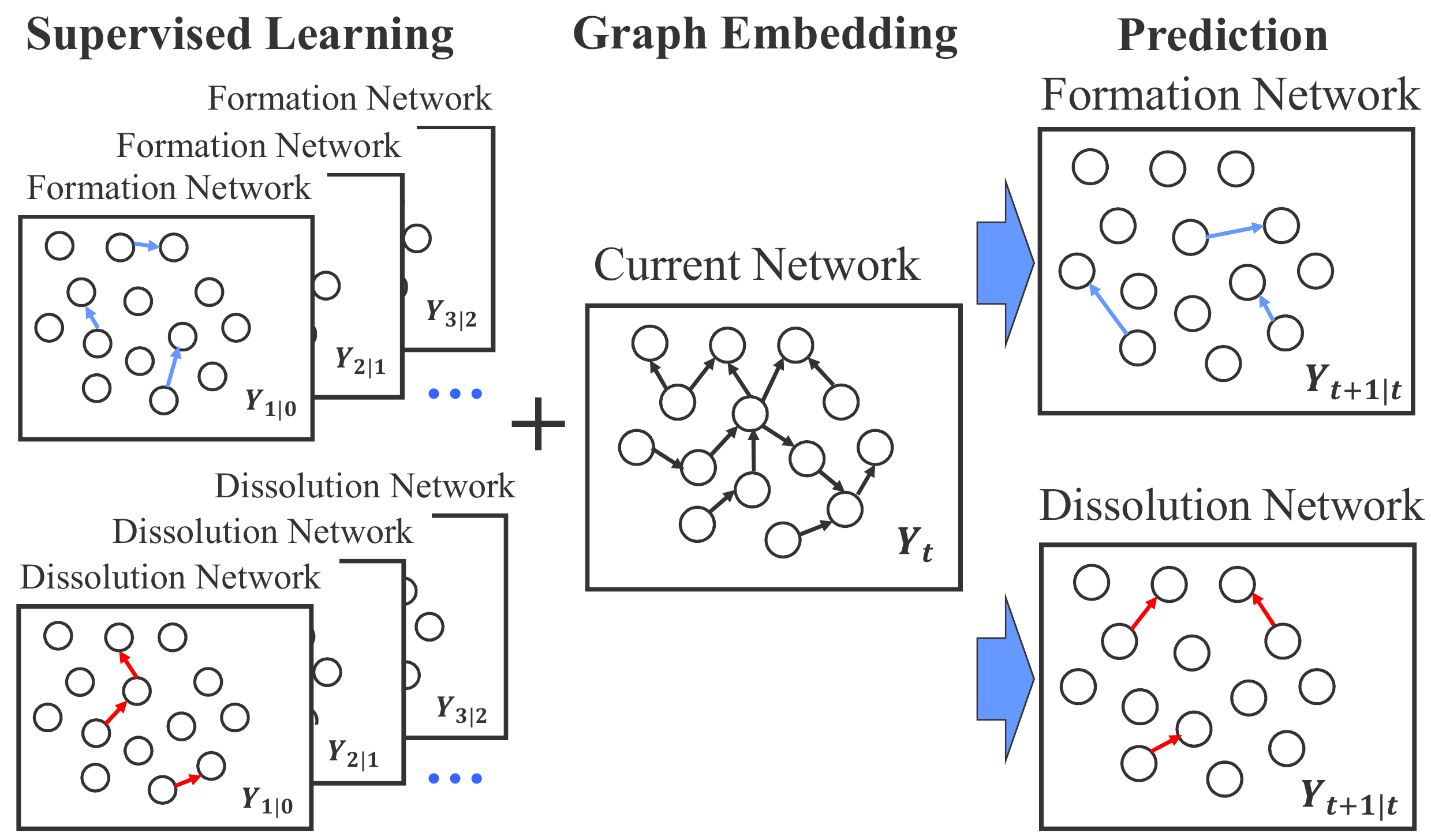} 
			\caption{Overview of our semi--supervised graph embedding approach to dynamic link prediction.} 
			\label{fig:1} 
		\end{center} 
	\end{figure*}

	To gain a better understanding of our model, we suggest the following intuitive interpretation (refer to Fig \ref{fig:1} for an overview of our approach).  While the temporal information concerning past link formation and link dissolution networks provides a direct target signal for which nodes were more likely to form or dissolve links with each other, these networks are usually much sparser than the current network.  Thus, by only using the past network information we may not have enough information to learn the complex--valued vector bilinear model sufficiently.  On the other hand, the current network can be seen as providing a different dimension, such as a spatial dimension in spatiotemporal modeling, which is independent of the temporal information.  Our strategy is to leverage this extra dimension to enhance the model learned from our supervised task.  Thus the power of graph embedding to effectively learn a distributional context capable of predicting nearby nodes is used in our model to force nearby nodes in the network to have similar complex-valued vectors \cite{Perozzi2014}.  We show that our semi--supervised approach gives better predictive performance than using a supervised or an unsupervised approach alone.

	The main contributions of this paper are as follows.
	
	\begin{itemize}
		\item We propose a simple and scalable discrete time semi--supervised graph embedding approach to dynamic link prediction capable of incorporating both temporal and cross--sectional network structures.
		\item Our model is one of the few approaches capable of learning different embeddings for both the formation and dissolution processes.
		\item Experiments with three real--world datasets show significant empirical improvements especially when predicting link dissolution.
	\end{itemize}
	
	The rest of the paper is organized as follows.  We present our proposed model in Section 2.  Our training methodology is presented in Section 3.  We give empirical results in Section 4, followed by related work and conclusions in Sections 5 and 6.

	\section{Proposed Method}
	
	We refer to our link prediction method as \textit{SemiGraph}, which has the objective functions in Eq. (2.9) and Eq. (2.10) for link formation and link dissolution, respectively.  Predictions are made using Eq. (2.13) and Eq. (2.14).

	\subsection{Notations}
	
	We now give a brief explanation of our notation and definitions for some terminology.  Consider a sequence of directed networks defined as a set of adjacency matrices $G=\{G_{1}, G_{2},\dots,G_{t}\}$, where $G_{ijt}$ equals $1$ if the link $i -> j$ exists at time $t$ and equals $0$ otherwise.  Let $V$ denote the set of nodes in the union of each snapshot of the network $G_{1} \cup G_{2} \cup \dots \cup G_{t}$, and let $|V|$ denote the number of nodes in the union of all the networks.  The goal of this paper is to predict the transition of the network from $G_{t}$ to $G_{t+1}$ using the information up to $G_{t}$.
	
	We define three kinds of network.  The \textit{current network} is the network state just before prediction.  With the above definitions, this is simply $G_{t}$.  The past \textit{formation networks} are defined by concatenating all the link formation adjacency matrices until time $t$.  The adjacency matrix describing the link formation network at time $t$ is defined as
	
	\[
	\begin{cases}
	F_{ijt} = 1 \;\; if \;\; G_{ijt} - G_{ijt-1} = 1 \\
	F_{ijt} = 0 \;\; otherwise.
	\end{cases}
	\]
	
	\noindent The past \textit{dissolution networks} are defined similarly, where the adjacency matrix describing the link dissolution network at time $t$ is defined as 
	
	\[
	\begin{cases}
	D_{ijt} = 1 \;\; if \;\; G_{ijt} - G_{ijt-1} = -1 \\
	D_{ijt} = 0 \;\; otherwise.
	\end{cases}
	\]

	\subsection{Learning from past formation and dissolution networks}
	
	We start with the supervised part, which consists of learning a complex--valued vector bilinear model with past link formation and link dissolution information playing the role of target values in the training data.  The complex--valued matrix of the node representations (i.e. $C^{|V| \times d}$, where $|V|$ denotes the number of nodes in the network and $d$ the dimension of the learned representations) are learned separately for link formation and link dissolution.  These are learned in an identical manner, and we focus on the link formation case.
	
	Formally, let $(i,j)$ be a set of links in the past formation networks.  The set of past formation networks is restricted to the information from link formation networks for a time window $F_{t}, F_{t-1}, …, F_{t-p}$.  The loss function can be written as

	\begin{eqnarray} 
	\Sigma_{i,j\in(i,j)} log p(j|i) = \Sigma_{i,j\in(i,j)}  (Re(\overline{v}_{fi}^{T} W_{f} v_{fj}) - \nonumber\\ 
	log \Sigma_{j' \in Ne} exp(Re(\overline{v}_{fi}^{T} W_{f} v_{fj'}))),
	\end{eqnarray}

	\noindent where $Ne$ is the set of all edges that did not form links with $i$ in the past formation networks, $W_{f}$ is a diagonal complex--valued matrix defining the scaling of the basis, $v_{fi}$ is the complex vector representation for node $i$ with dimension $d$, $\overline{v}$ denotes the conjugate of $v$ (i.e. $\overline{v} = Re(v) - iIm(v)$) and Re() is a function keeping only the real part of a complex value.  The use of a complex--valued vector instead of a real--valued vector is to take into account symmetric as well as antisymmetric relations in both linear space and time complexity by using the Hermitian dot product \cite{Trouillon2016}
	
	\begin{eqnarray} 
	<u,v> = \overline{u}^T v,
	\end{eqnarray}
	
	\noindent where $u$ and $v$ are complex--valued vectors.  The Hermitian dot product has the nice property that $<u,v>$ does not necessarily equal $<v,u>$, making it possible to consider antisymmetric relations \cite{Trouillon2016}.  We also restrict each diagonal element of $W_{f}$ and $W_{d}$ to have an absolute value of 1 to make the model identifiable.
	
	It is often intractable to directly optimize Eq. (1) due to the normalization constant, and we use negative sampling to address this issue \cite{Mikolov2013}.  Formally, given a triple $(i,j,\gamma_{f})$, where $i$ and $j$ are nodes (we assume that $i \neq j$) and $\gamma_{f}$ is a binary label indicating whether a node pair exists in the past link formation networks (this is positive when links exists in the formation networks), we minimize the cross entropy loss of classifying the pair $i,j$ with a binary label $\gamma_{f}$:

	\begin{eqnarray} 
	I(\gamma_{f}=1) log \sigma(Re(\overline{v}_{fi}^{T} W_{f} v_{fj})) + \nonumber\\ I(\gamma_{f}=-1)log \sigma(-Re(\overline{v}_{fi}^{T} W_{f} v_{fj})),
	\end{eqnarray}
	
	\noindent where $I(.)$ is an indicator function that outputs $1$ when the argument is true and $0$ otherwise and $\sigma$ is a sigmoid function defined as $\sigma(x) = 1/(1 + e^{-x})$.  Therefore, the supervised loss with negative sampling can be written more succinctly as 
	
	\begin{eqnarray} 
	L_{fs} = E_{i,j,\gamma_{f}} log \sigma(\gamma_{f} Re(\overline{v}_{fi}^T W_{f} v_{fj})).
	\end{eqnarray}
	
	The supervised loss for past dissolution networks is defined in an identical manner, resulting in
	
	\begin{eqnarray} 
	L_{ds} = E_{i,j,\gamma_{d}} log \sigma(\gamma_{d} Re(\overline{v}_{di}^T W_{d} v_{dj})).
	\end{eqnarray}

	\subsection{Graph Embedding from the Current Network}
	
	The unsupervised part of our model consists of a graph embedding defined by the current network.  In previous works, a Skipgram model \cite{Mikolov2013} is used to learn the embedding and we adhere to this approach.  Given a pair of an instance and its context (i.e. $(i,c)$), the loss function can be written as 
	
	\begin{eqnarray} 
	\Sigma_{i,c \in (i,c)} log p(c|i) = \Sigma_{i,c \in (i,c)} (Re(\overline{v}_{fi}^{T} u_{fc}) - \nonumber\\ log \Sigma_{j’ \in Ne} exp(Re(\overline{v}_{fi}^{T} u_{fc}))),
	\end{eqnarray}

	\noindent where $v_{fi}$ is the complex vector representation for node $i$ as used in Eq. (1) and $u_{fc}$ is a parameter for the Skipgram model.  A context for each node is generated by performing a truncated random walk (i.e. deep walk) starting from the instance node \cite{Perozzi2014}.  Although other types of walk besides the simple random walk (such as a breadth--first walk) are possible \cite{Tang2015}, preliminary experiments showed that the difference is marginal and we use the simple deep walk in this paper.  As in Eq. (1), Eq. (6) is intractable due to the normalization constants and we again resort to negative sampling, resulting in 
	
	\begin{eqnarray} 
	L_{fu} = E_{i,c,\gamma_{c}} log \sigma(\gamma_{c} Re(\overline{v}_{fi}^T u_{fc})).
	\end{eqnarray}
	
	The unsupervised loss for link dissolution is developed in an identical manner, resulting in 
	
	\begin{eqnarray} 
	L_{du} = E_{i,c,\gamma_{c}} log \sigma(\gamma_{c} Re(\overline{v}_{di}^T u_{dc})).
	\end{eqnarray}

	\subsection{Semi--supervised Graph Embedding Approach}
	
	Given the loss functions defined in the previous sections, the loss functions for our framework can be expressed as 
	
	\begin{eqnarray} 
	L_{f} = L_{fs} + \lambda_{f} L_{fu} 
	\end{eqnarray}
	
	\noindent for learning link formation and
	
	\begin{eqnarray} 
	L_{d} = L_{ds} +\lambda_{d} L_{du} 
	\end{eqnarray}
	
	\noindent for learning link dissolution.  The $L_{fs}$ and $L_{ds}$ terms are the supervised losses for predicting past formation or dissolution networks, respectively, and $L_{fu}$ and $L_{du}$ are the unsupervised losses for predicting the graph context from the current network.  The loss function is similar in spirit to graph--based semi--supervised learning \cite{Zhou2004,Zhu2003}, where graph embedding was used instead of the graph Laplacian as in \cite{Yang2016}.

	\subsection{Prediction}
	
	Prediction is made by using the learned complex--valued vectors and matrices $v_{f}$, $v_{d}$, $W_{f}$ and $W_{d}$.  A straightforward approach is to predict
	
	\begin{eqnarray} 
	p(G_{ijt+1}=1| G_{ijt}=0) = \nonumber\\ \sigma(Re(\overline{v}_{fi}^{T} W_{f} v_{fj}))
	\end{eqnarray}
	
	\noindent for link formation and 
	
	\begin{eqnarray} 
	p(G_{ijt+1}=0| G_{ijt}=1) = \nonumber\\
	\sigma(Re(\overline{v}_{di}^{T} W_{d} v_{dj}))
	\end{eqnarray}
	
	\noindent for link dissolution.  Although this simple prediction works quite well in practice, the predictive performance can be further improved by combining the predictions as
	
	\begin{eqnarray} 
	p(G_{ijt+1}=1| G_{ijt}=0) = \nonumber\\ \sigma(Re(\overline{v}_{fi}^{T} W_{f} v_{fj})) +Re(\overline{v}_{di}^{T} W_{d} v_{dj}))
	\end{eqnarray}
	
	\noindent for link formation and 
	
	\begin{eqnarray} 
	p(G_{ijt+1}=0| G_{ijt}=1) =  \nonumber\\ \sigma(Re(\overline{v}_{di}^{T} W_{d} v_{dj})) + Re(\overline{v}_{fi}^{T} W_{f} v_{fj}))
	\end{eqnarray}
	
	\noindent for link dissolution.  The underlying understanding of this prediction is that link formation and link dissolution are more likely to be driven by a rewiring process:  Thus the more likely a node is to form new links, the more likely the node is to dissolve an existing link at the same time.  Although subtracting the two effects, as in
	
	\begin{eqnarray} 
	p(G_{ijt+1}=1| G_{ijt}=0) =  \nonumber\\ \sigma(Re(\overline{v}_{fi}^{T} W_{f} v_{fj})) - Re(\overline{v}_{di}^{T} W_{d} v_{dj}))
	\end{eqnarray}
	
	\noindent for link formation and 
	
	\begin{eqnarray} 
	p(G_{ijt+1}=0| G_{ijt}=1) =  \nonumber\\ \sigma(Re(\overline{v}_{di}^{T} W_{d} v_{dj})) - Re(\overline{v}_{fi}^{T} W_{f} v_{fj}))
	\end{eqnarray}
	
	\noindent for link dissolution, is also reasonable (i.e. a growing network where the more likely a node is to form links the less likely the node is to lose a link), in our experiments Eqs. (2.13) and (2.14) outperform the other prediction method, so we use this prediction in our experiments.

	\section{Training}
	
	We use stochastic gradient descent to train our model \cite{Bottou2010}.  We first sample a node and perform a deep walk \cite{Perozzi2014} to sample the context nodes from a network.  We then sample negative samples from the current network, past formation networks, and past dissolution networks.  Equipped with these positive and negative samples, we take a gradient step with learning rate $\eta_1$ for $v_f$, $v_d$, $u_f$ and $u_d$.  
	
	Each diagonal element of $W_{f}$ and $W_{d}$ is learned in a different manner.  As noted before, to make the model identifiable we restrict each diagonal element of $W_{f}$ and $W_{d}$ to take an absolute value of $1$.  Thus each diagonal element of $W_{f}$ can be rewritten as
	
	\begin{eqnarray} 
	W_{f}(i,i) = cos(\theta) + isin(\theta) 
	\end{eqnarray}
	
	\noindent for $i = 1,2,…,d$.  We take a gradient step with learning rate $\eta_{2}$ in $\theta$ instead.  All the off--diagonal elements are set to $0$.


	\section{Experiments}
	
	Our empirical investigations are based on three real--world networks: a world trade network, an interfirm buyer--seller network and bipartite customs data between Japan and the US (Japan to US exports only).  
	
	\subsection{Data}	
	
	We next give a brief outline of the data used.
	\begin{itemize}
		\item WorldTrade is a network of world trade relationships among 50 countries from 1981 to 2000 \cite{Westveld2011}.  We define two countries to be linked if the trading volume was above the 90th percentile for all trade in a given year.
		\item FirmNetwork is an interfirm buyer--seller network for Japan from 2003 to 2012.  We use a subset of this dataset, restricting our attention to firms in Hokkaido in the northern part of Japan \cite{Hisano2015}.
		\item Customs is a bipartite network dataset that records the names of exporters and consignees of trade from Japan to the US.  The data was obtained from the US customs office and covers the period from January 2003 to December 2014.  We focus on firms that had more than 500 transactions during the time period, which results in 431 Japanese firms and 603 US firms.  To adjust for seasonal effects, we aggregate the network data on a yearly basis resulting in snapshots of 12 networks.  Two firms are linked if there was a trade relation more than once a year.
	\end{itemize}
	
	\noindent The basic statistics for each dataset are reported in Table 1.

	\begin{table*}[!htbp] \centering  
		\label{table:1}  
		\begin{tabular}{lcccccc}
			\\[-1.8ex]\hline  
			\hline \\[-1.8ex]  
			Dataset & \multicolumn{1}{c}{Num Nodes} & \multicolumn{1}{c}{Num Edges} & \multicolumn{1}{c}{Num Unique Edges} & \multicolumn{1}{c}{Ave Form} & \multicolumn{1}{c}{Ave Diss} & \multicolumn{1}{c}{Snapshots} \\  
			\hline \\[-1.8ex]  
			WorldTrade & 50 & 6620 & 477 & 16.7 & 16.7 & 20 \\ 
			Firm & 690 & 13108 & 1995 & 118.9 & 126.3 & 10  \\ 
			Customs & 1043 & 7825 & 1488 & 113.9 & 126 & 12 \\ 
			\hline \\[-1.8ex]  
		\end{tabular} 
		\caption{Statistics for datasets.  Num Edges denotes the total number of interactions, Num Unique Edges denotes the number of distinct interactions, Ave Form denotes the average number of formed edges, Ave Diss denotes the average number of dissolved edges and Snapshots denotes the number of discrete time points observed in our datasets.}  
	\end{table*}

	\subsection{Evaluation Criteria}
	
	Given a training network $G_{1:t}$, we predict the transition from time $t$ to time $t+1$ which consists of a link formation network (i.e. $F_{t+1}$) and a link dissolution network (i.e. $D_{t+1}$) as shown in Fig ~\ref{fig:1}.  For link prediction accuracy, we use the area under the receiver operating characteristic curve (AUC), where the value is calculated for both link dissolution networks and link formation networks.  The AUC has the nice property that it is not influenced by the distribution of classes, making it suitable in our setting where classes (e.g. formed or not formed, dissolved or not dissolved) are highly imbalanced \cite{Menon2011}.  Higher AUC values indicate better link prediction performance.

	\subsection{Baseline Methods}
	
	We compare our prediction algorithm with the following baselines.  
	
	\begin{itemize}
		\item Adamic-Adar (AA): scores are calculated as the weighted variation of common neighbors \cite{AdaAda2003} using the current network only.  
		\item Preferential attachment (PA): scores are calculated as the product of the degree of each node from the current network.
		\item Last time of linkage (LL): scores are calculated by ranking pairs in ascending order according to the last time of linkage \cite{Tylenda2009}.
	\end{itemize}
	
	We also compute AA-all and PA-all, which are computed over the union of all networks until the current network.  The graph heuristic approaches presented here are simple but have been shown to be surprisingly hard to beat in practice, making them good baselines for comparison \cite{Liben-Nowell2003,Tylenda2009,Sarkar2014}.  In particular, LL has been shown to often be among the best heuristic measures for link prediction \cite{Tylenda2009, Sarkar2014}.  When predicting link dissolution, we use the complementary score method as in \cite{Preusse2013,Preusse2014}.  We also compare our model with unsupervised graph embedding and supervised approach (i.e. our model without the graph embedding term) to clarify the improvement in semi--supervised learning.  Throughout all of the experiments, we set $d=3$, the number of walks as five, $\lambda_{f}=\lambda_{d}=0.05$, $\eta_{1}=0.05$, $\eta_{2}=5 \times 10^-6$ and $p=t-1$ (i.e. using all past information).  The learning rate is decreased linearly with the number of nodes that have been used for training to that point.

	\subsection{Experimental Results}
	
	Results for the link formation prediction task are presented in Table 2.  We make the following observations.  For the Firm and Customs datasets, our proposed method is the best, but for the WorldTrade dataset, PA-all shows slightly better performance than our method.  Nonetheless, for all the networks studied here, our proposed method is among the top performing methods.  We observe that the state of the art baseline methods work quite well especially when using the union of past networks.  For the bipartite Customs dataset, AA and AA-all perform almost as the same as random selection because we do not have enough linkage information to calculate common neighbors.  Our method also shows significant improvements over graph embedding and supervised learning.  In this experiment, supervised learning is outperformed by our method by around 15 \% - 18 \%, while graph embedding is outperformed by more than 40 \%, suggesting the added value of our semi--supervised approach.

	\begin{table}[!htbp] \centering  
		\label{table:1}  
		\begin{tabular}{lcccccccc}
			\\[-1.8ex]\hline  
			\hline \\[-1.8ex]  
			Dataset & \multicolumn{1}{c}{WorldTrade} & \multicolumn{1}{c}{Firm} & \multicolumn{1}{c}{Customs}  \\  
			\hline \\[-1.8ex]  
			AA & 0.647 & 0.615 & 0.5 \\ 
			PA & 0.761 & 0.709 & 0.517 \\ 
			AA-all & 0.643 & 0.689 & 0.5 \\ 
			PA-all & \bf{0.885} & 0.787 & 0.748 \\
			LastTime & 0.762 & 0.778 & 0.834 \\
			Supervised & 0.703 & 0.717 & 0.764 \\
			GraphEmb & 0.588 & 0.581 & 0.606 \\
			SemiGraph & 0.835 & \bf{0.828} & \bf{0.842}  \\
			\hline \\[-1.8ex]  
		\end{tabular} 
		\caption{AUC for link formation prediction}  
	\end{table}

	Results for link dissolution prediction are presented in Table 3.  We make the following observations.  For all the experiments our method performs better than the state of the art baseline methods.  It is worth noting that our method outperforms the other methods quite significantly for the Firm dataset, whereas other unsupervised approaches show almost no signs of predictability.  In this experiment, supervised learning is outperformed by our method by around 7 \% - 13 \%, suggesting again the added value of our semi--supervised approach.  The graph embedding approach shows almost no sign of predictability in predicting link dissolution.  We also observe that when predicting link dissolution, adding past information does not necessarily increase the predictive performance.  For the Customs dataset, using the complementary score does not necessarily improve predictability, and a better AUC score can be obtained by using the normal PA score.

	\begin{table}[!htbp] \centering  
		\label{table:1}  
		\begin{tabular}{lcccccccc}
			\\[-1.8ex]\hline  
			\hline \\[-1.8ex]  
			Dataset & \multicolumn{1}{c}{WorldTrade} & \multicolumn{1}{c}{Firm} & \multicolumn{1}{c}{Customs}  \\  
			\hline \\[-1.8ex]  
			AA & 0.638 & 0.522 & 0.496 \\ 
			PA & 0.711 & 0.504 & 0.325 \\ 
			AA-all & 0.642 & 0.488 & 0.49 \\ 
			PA-all & 0.629 & 0.458 & 0.467 \\
			LastTime & 0.596 & 0.529 & 0.671 \\
			Supervised & 0.651 & 0.674 & 0.620 \\
			GraphEmb & 0.486 & 0.514 & 0.395 \\
			SemiGraph & \bf{0.737} & \bf{0.725} & \bf{0.684}  \\
			\hline \\[-1.8ex]  
		\end{tabular} 
		\caption{AUC for link dissolution prediction.}  
	\end{table}

	To see how an increase in past information affects the performance of our proposed model, we report results on predicting the transition of a network for the years 2005 to 2012 for the Firm dataset.  Because we only have ten snapshots of the network, the prediction in 2005 is based on only one past transition and the last network before prediction.  We observe that for link formation prediction, almost all the methods including our proposed method show improved accuracy with an increase in past information.  Our method is among the best performing methods, with a performance comparable to PA-all.  Comparing our performance with supervised learning (our method without graph embedding), we clearly see the benefit of our semi--supervised approach.  For link dissolution, we observe that our method performs better than the baseline methods.  Although supervised learning sometimes performs slightly better than our method, overall we observe the added value of our semi--supervised approach.  Although less clear than link formation prediction we also observe that our method show improved accuracy with an increase in past information.

	\begin{figure*}[!h]
		\begin{minipage}[b]{0.495\linewidth}
			\centering
			\includegraphics[keepaspectratio, scale=0.4]
			{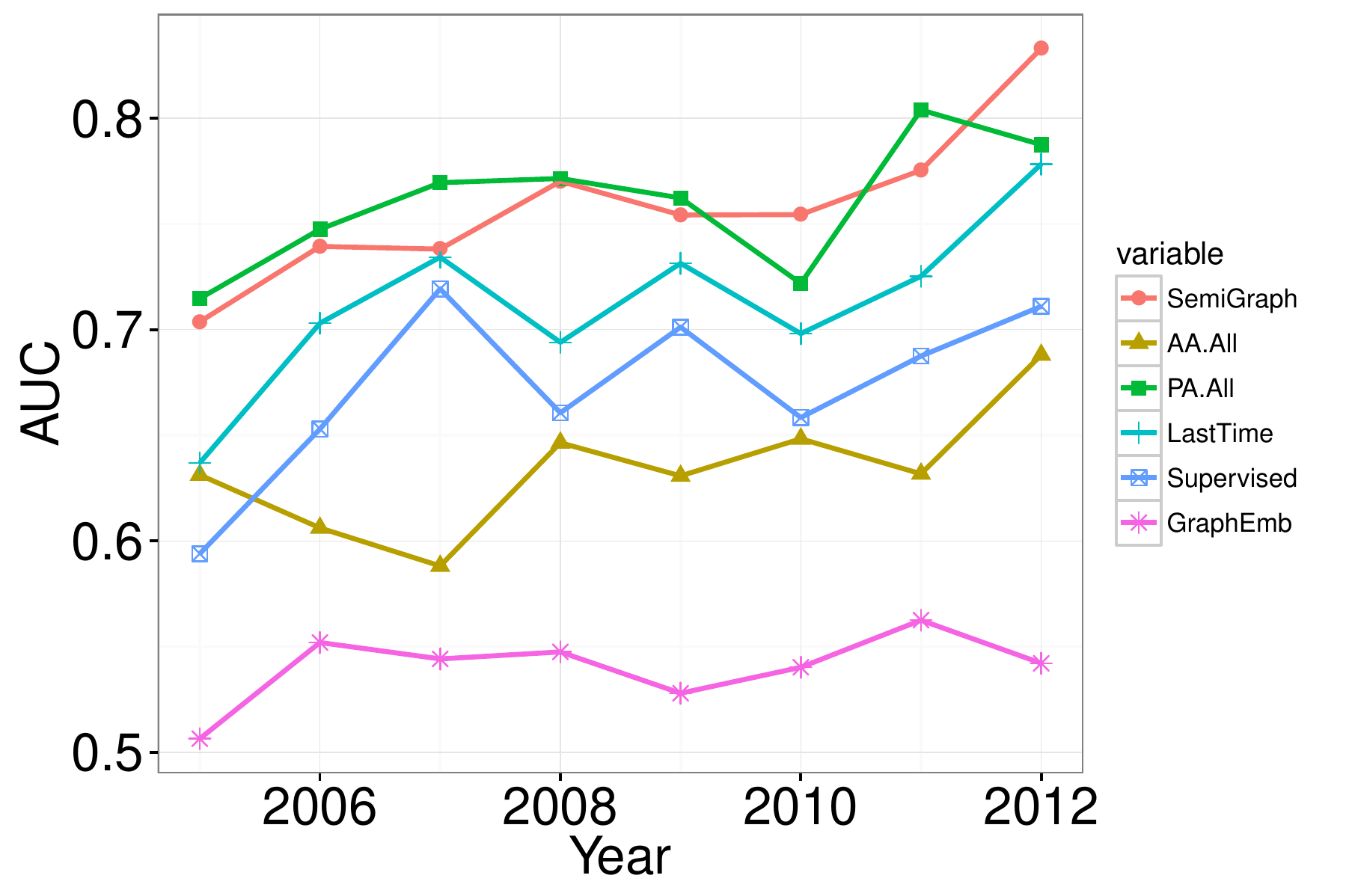}
			\subcaption{Link formation}
			\label{fig_pos_2003}
		\end{minipage}
		\begin{minipage}[b]{0.495\linewidth}
			\centering
			\includegraphics[keepaspectratio, scale=0.4]
			{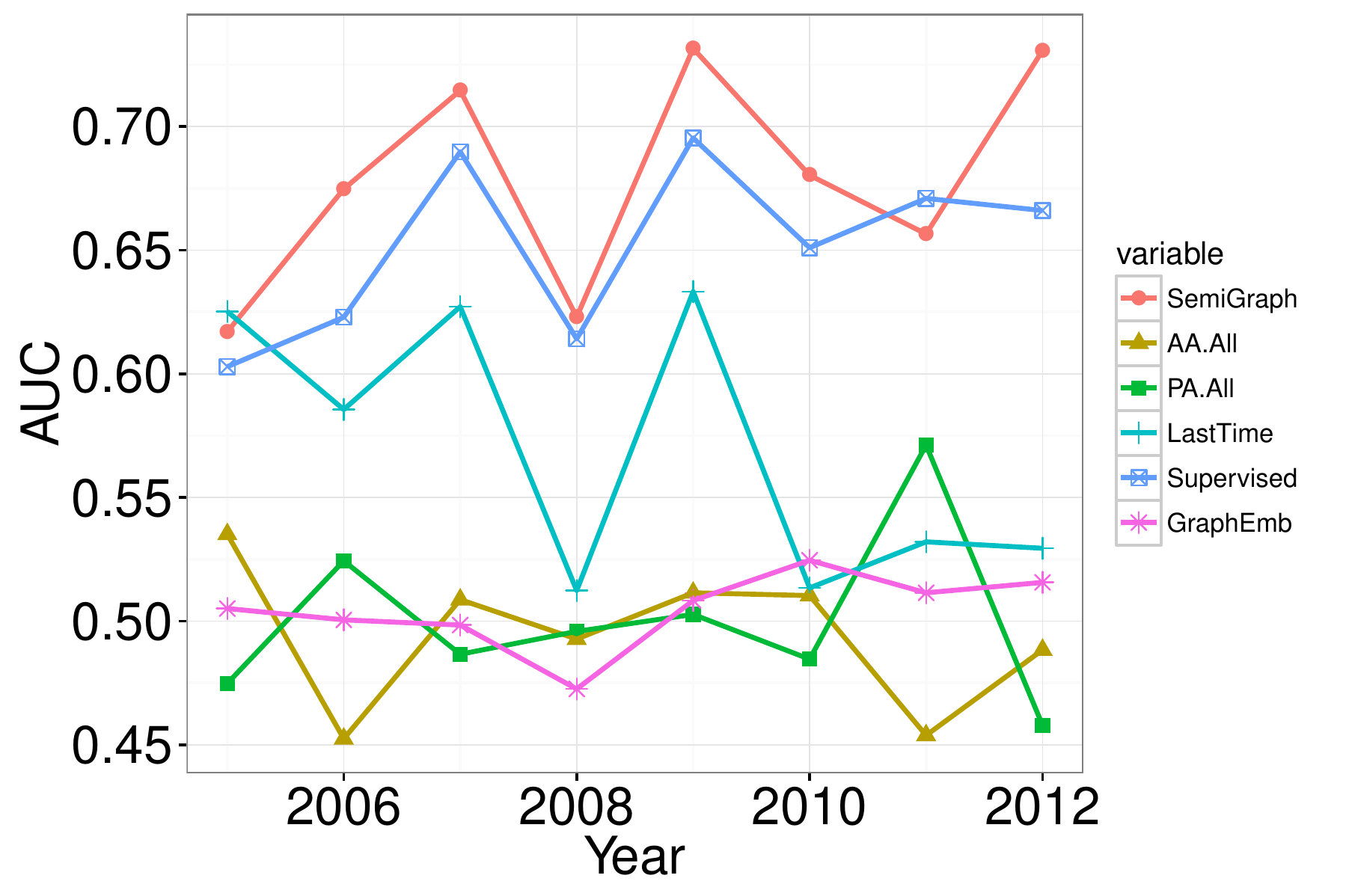}
			\subcaption{Link dissolution}
			\label{fig_pos_2004}
		\end{minipage}
		\caption{AUC for link formation and link dissolution prediction for the Firm dataset.}
		\label{fig5s}
	\end{figure*}

	\subsection{Parameter Sensitivity}
	To evaluate how changes to the parametrization affects the final predictive performance, we report the effect of varying the number of dimensions and $\lambda$ (we set $\lambda := \lambda_{f} = \lambda_{d}$).  Other parameters are held fixed as before.  Figure 3(a) shows the effect of varying the number of dimensions, and shows that while the performance does not vary greatly, the optimum seems to be three.  Figure 3(b) examines the effect of varying $\lambda$.  This shows a clear improvement compared to supervised learning (i.e. $\lambda=0$), where the optimum value seems to be around 0.05.  Beyond that, the performance gradually deteriorates as $\lambda$ increases.  These experiments show that although the usefulness of our model depends on several parameters, the choice is not too sensitive to these parameters.

	\begin{figure}[!h]
		\begin{minipage}[b]{0.495\linewidth}
			\centering
			\includegraphics[keepaspectratio, scale=0.25]
			{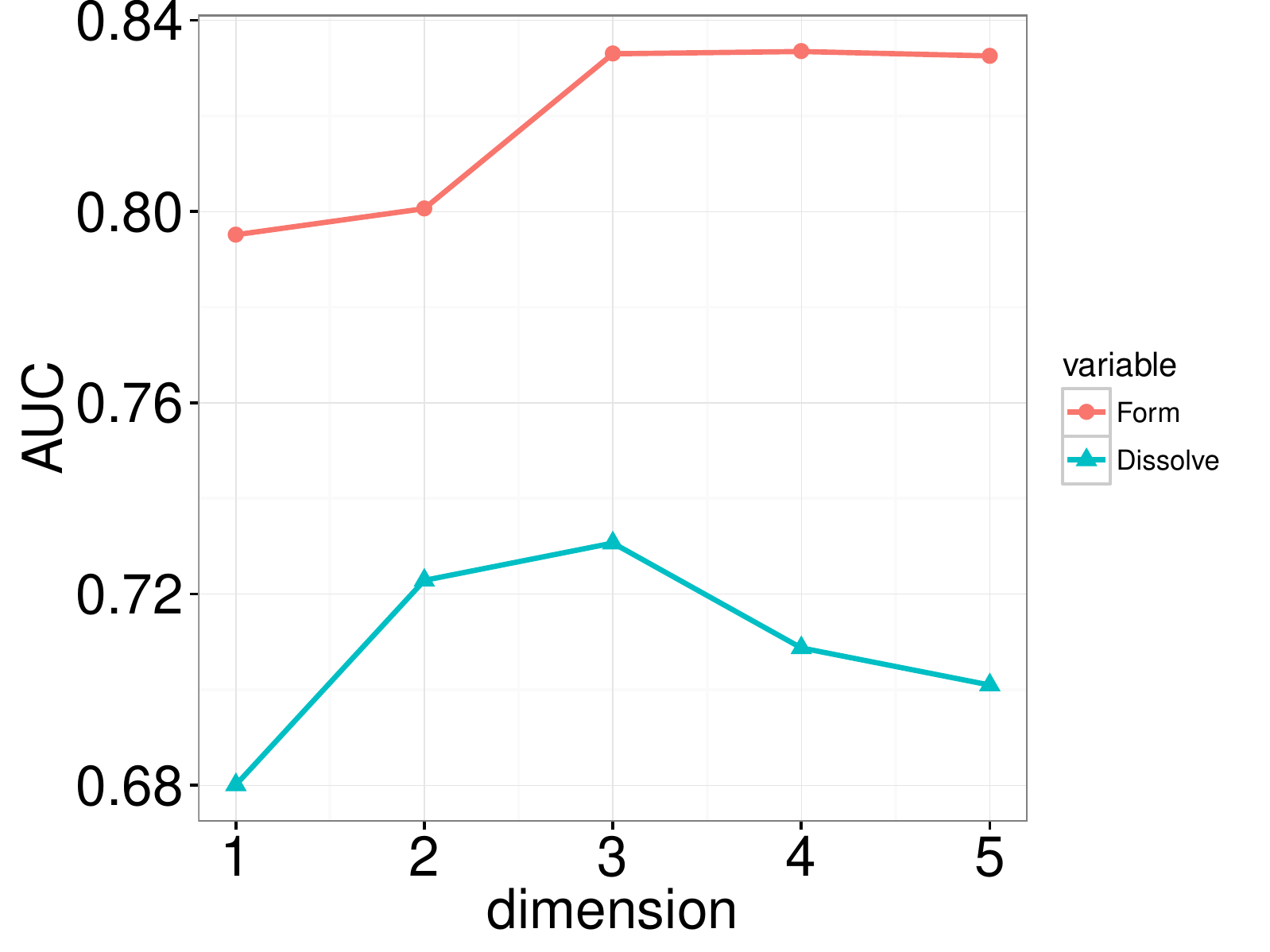}
			\subcaption{Stability over dimension $d$.}
			\label{fig_pos_2003}
		\end{minipage}
		\begin{minipage}[b]{0.495\linewidth}
			\centering
			\includegraphics[keepaspectratio, scale=0.25]
			{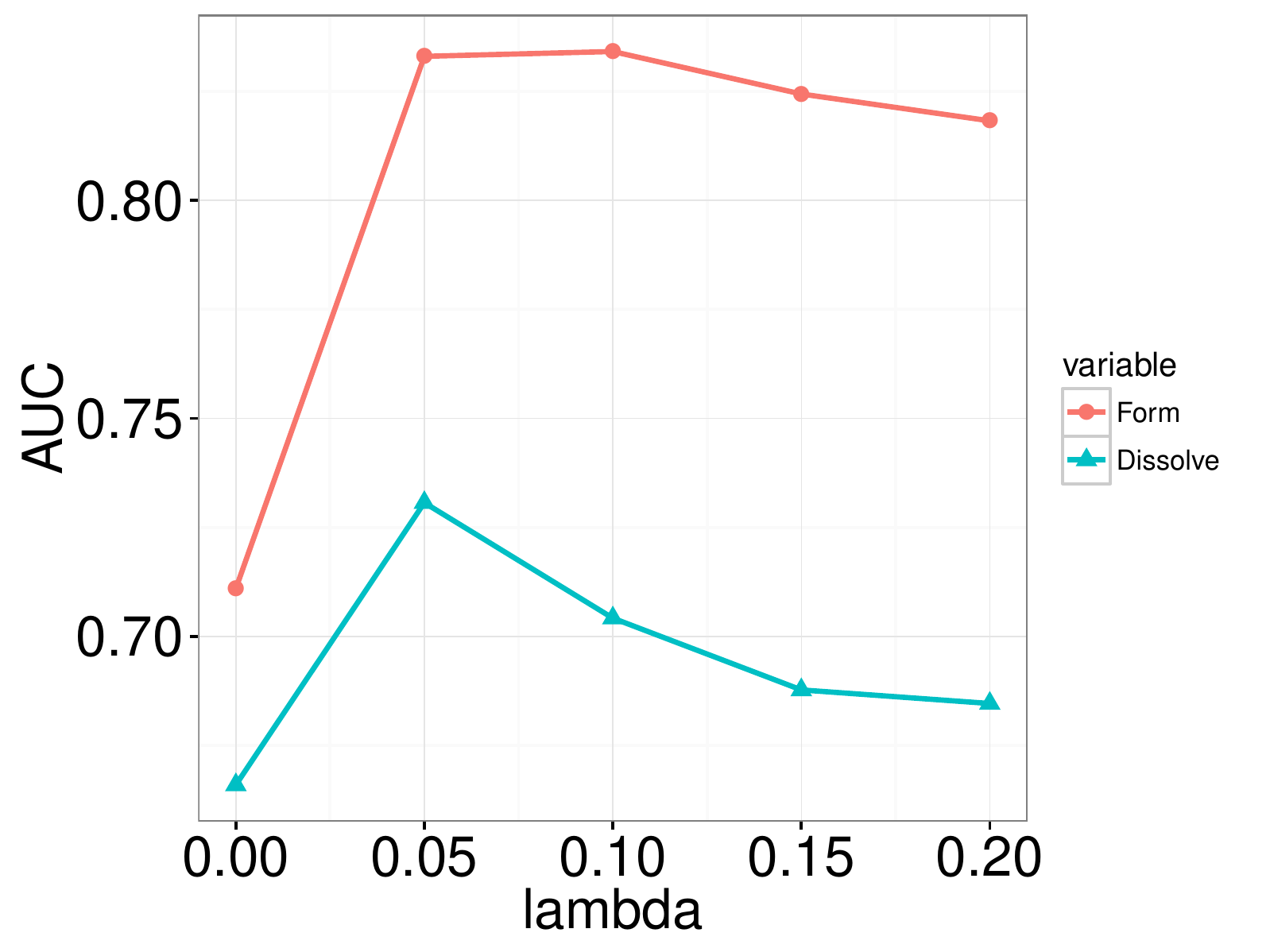}
			\subcaption{Stability over $\lambda$.}
			\label{fig_pos_2004}
		\end{minipage}
		\caption{Parameter sensitivity.}
		\label{fig5s}
	\end{figure}

	\section{Related work}		
	\subsection{Link Prediction}
	
	The static link prediction problem has been extensively studied in the literature \cite{Liben-Nowell2003}.  Among the many proposed approaches, graph--based heuristics are the most popular due to their simplicity and high performance on a variety of practical problems \cite{AdaAda2003}.  In the dynamic setting, \cite{Tylenda2009,Sarkar2014} examined extensions of existing static graph--based heuristic measures for temporal link prediction.  They showed that extremely simple graph--based heuristic measures such as last time to link work surprisingly well in practice.  
	
	\subsection{Link Dissolution Prediction}
	
	Previous research focusing on predicting link dissolution is much less common than for link formation prediction.  Recent research includes \cite{Kwak2012}, which studied unfollowing behavior on twitter, \cite{Yang2013} which studied unfriending behavior on Facebook and \cite{Preusse2013,Preusse2014} which studied link dissolution on Wikipedia.  In all of these previous studies, it was shown that predicting link dissolution is harder than predicting link formation.  Compared to these approaches, where information additional to network information is required to perform prediction, our approach is versatile in the sense that we only need snapshots of network information.

	\subsection{Other Related Approaches}
	
	From a supervised learning perspective, our approach can be seen as a descendant of a latent feature or matrix factorization approach to link prediction \cite{Menon2011,Kolar2010}.  The main differences are 1) learning past link formation and dissolution dynamics directly as well as separately, 2) using complex--valued vectors to make it possible take into account symmetric as well as antisymmetric relations for both linear space and time complexity and 3) the unsupervised graph embedding part proposed in this paper.  Bayesian extensions of latent feature models also exist \cite{Hoff2003}, with some studies allowing for an infinite number of latent features \cite{Miller2009}.

	Semi--supervised approaches to dynamic link prediction have also previously been explored.  In \cite{Kashima2009,Raymond2010}, Link Propagation was proposed, where a kernel--based semi--supervised approach to link prediction is performed by constructing a kernel that compares node pairs that constrains the values in the adjacency matrix to vary smoothly according to the kernel.  Our approach is arguably simpler than their approach, as the effectiveness of their method depends on the choice of kernel which has to be pre-specified.

	A popular approach to temporal link prediction is based on extensions of static latent space models \cite{Sarkar2005,Sewell2015} and mixed membership stochastic block models \cite{Fu2009,Xing2010} in a temporal setting.  The main idea is to model longitudinal network data as smooth trajectories in a latent space.  In social networks, several models extending the exponential random graph models to a dynamic setting have been proposed \cite{Guo2007,Krivitsky2014}.  Along these lines, \cite{Krivitsky2014} is a nice extension of the exponential random graph models that enables different modeling for both link formation and link dissolution dynamics.  A model similar to the exponential random graph model was also proposed for statistical relational learning \cite{Taskar2003}.  However, these approaches are generally computationally expensive which limits scalability.  Other studies concerning temporal networks include \cite{Westveld2011}, which proposed a longitudinal mixed effect model capable of learning latent representations that evolves in a simple auto-regressive manner, \cite{Emile2010} where a vector autoregressive model was used for link prediction in dynamic graphs and \cite{Dunlavy2011} which proposed a tensor--based method to predict periodic temporal data with multiple patterns.   
	

	\section{Conclusions}
	
	We have proposed \textit{SemiGraph}, a simple discrete--time semi--supervised graph embedding approach to link prediction in dynamic networks.  Our model is capable of learning different embeddings for both formation and dissolution dynamics.  To show the effectiveness of our approach, we focused on predicting the \textit{transition} of a network, including both link formation prediction and link dissolution prediction.  We have showed that our method outperforms previous state of the art baseline methods in predicting link dissolution and is comparable to state of the art methods in predicting link formation through experiments using a variety of real--world networks.

	\bibliography{Hisano_myref} 
	\bibliographystyle{ieeetr}
	
\end{document}